\documentclass[accepted]{uai2026} 
                        
\usepackage[american]{babel}

\usepackage{amsmath}
\usepackage{amsfonts}
\usepackage{amssymb}
\usepackage{mathtools}
\usepackage{nicefrac}
\usepackage{amsthm}
\usepackage{thmtools}
\usepackage{thm-restate}

\usepackage[nameinlink,capitalize,noabbrev]{cleveref}

\usepackage{natbib} 
\usepackage{booktabs} 
\usepackage{tikz} 
\usepackage{subcaption} 

\title{A Compression Perspective on Simplicity Bias}

\author[1,2]{\href{mailto:tom.marty@mila.quebec}{Tom Marty}$^*$}
\author[1,2]{Eric Elmoznino}
\author[1,2]{Leo Gagnon}
\author[1,2]{Tejas Kasetty}
\author[1,2]{Mizu Nishikawa-Toomey}
\author[1,2]{\\Sarthak Mittal}
\author[1,2]{Guillaume Lajoie}
\author[1,2]{\href{mailto:dhanya.sridhar@mila.quebec}{Dhanya Sridhar}$^\dagger$}

\affil[1]{Mila -- Quebec AI Institute}
\affil[2]{Universit\'e de Montr\'eal}
\begin{document}

\maketitle

\begin{abstract}
    Deep neural networks exhibit a simplicity bias, a well-documented tendency to favor simple functions over complex ones. In this work, we cast new light on this phenomenon through the lens of the Minimum Description Length principle, formalizing supervised learning as a problem of optimal two-part lossless compression. Our theory explains how simplicity bias governs feature selection in neural networks through a fundamental trade-off between model complexity (the cost of describing the hypothesis) and predictive power (the cost of describing the data). Our framework predicts that as the amount of available training data increases, learners transition through qualitatively different features -- from simple spurious shortcuts to complex features -- only when the reduction in data encoding cost justifies the increased model complexity. Consequently, we identify distinct data regimes where increasing data promotes robustness by ruling out trivial shortcuts, and conversely, regimes where limiting data can act as a form of complexity-based regularization, preventing the learning of unreliable complex environmental cues. We validate our theory on a semi-synthetic benchmark showing that the feature selection of neural networks follows the same trajectory of solutions as optimal two-part compressors.
\end{abstract}

\section{Introduction}
\label{sec:intro}
Several works point to the simplicity bias exhibited in the training of deep neural networks \citep{arpit2017closer,valle2018deep,shah2020pitfalls,mingard2025deep}. 
Put simply, simplicity bias refers to the tendency of learning algorithms such as stochastic gradient descent (SGD) to find solutions that encode ``simple functions.''
In this paper, we take the view that the notion of simplicity bias is captured by the Minimum Description Length (MDL) principle \citep{grunwald_tutorial_2004,grunwald2007minimum}. The MDL principle formalizes simplicity from the lens of lossless compression: we can describe a dataset with as few bits as possible if we have a model that accurately predicts the data (i.e., minimizes the negative log-likelihood) while requiring minimal bits to encode the model itself (i.e., by leveraging the structure in the data). In this paper, we revisit simplicity bias in neural networks by studying whether optimal compression serves as a predictive theory of neural network behavior. We then characterize the resulting implications of this theory for out-of-distribution (OOD) generalization.

A large body of work on simplicity bias in neural networks has focused on supervised learning in the presence of \emph{spurious features}: latent properties of inputs that are easy to extract and use for prediction, but are unreliable. For example, consider the problem of classifying images of birds as ``water-based'' or ``land-based'' as introduced in \cite{sagawa2020}. The background of the image (the presence or lack of water) often co-occurs with its corresponding label, offering an appealing shortcut that a machine learning classifier can easily exploit to achieve good performance in-distribution, but which leads to poor predictions if the spurious co-occurrence disappears in a shifted test distribution. In contrast, a human labeller might rely exclusively on a bird's phenotypical characteristics, classifying images accurately even when a water bird is on land.
Studies point to the fact that when learning predictive models from a static dataset, a preference for simple functions can be a curse, leading to models with poor generalization under distribution shift due to their reliance on spurious features \citep{geirhos2020shortcut, teney2022evading, vasudeva2023mitigating}.

This paper contributes to our understanding of the interactions between learning under simplicity bias and OOD generalization. Concretely, we focus our MDL predictive theory on heterogeneous data sampled from a mixture of underlying distributions that we refer to as environments. 
We vary relevant properties of heterogeneous data -- the amount of training data available, the complexity and predictiveness of different features, etc. -- and use our theory to predict which features a learned model will use in each data regime. 

We summarize the following contributions:
\begin{itemize}
    \item  We formalize supervised learning under simplicity bias as two-part lossless compression, following the MDL principle. We operationalize our predictive theory using prequential coding: a tractable method for approximating the complexity of a function.
    \item We show how simplicity bias gives rise to a dynamic feature preference that \textbf{depends on the amount of training data available}. Under this compression paradigm, learners behave as MDL-optimal compressors, shifting between distinct solutions when the reduction in data encoding cost outweighs the increase in model complexity.
    \item We provide empirical evidence on heterogeneous semi-synthetic datasets demonstrating that our MDL-based framework accurately predicts the OOD behavior of learners, validating it as a quantitative theory for predicting generalization failure modes in limited data regimes.
\end{itemize}
Code to reproduce our experiments is available at https://github.com/3rdCore/complicity.

\section{Theory}
\label{sec:theory}

In this section, we formalize the connection between simplicity bias and data compression. We first cast supervised learning as a two-part lossless compression problem in \cref{subsec:two_part}, where the learner needs to jointly minimize training error and model complexity. We then analyze in \cref{subsec:analysis} how the MDL-optimal solution shifts across data regimes, showing that the dominant hypothesis transitions from simple, low-cost solutions to more predictive but complex ones as the amount of training data grows. Finally, in \cref{subsec:robust}, we focus this analysis on the robust learning setting, where we identify two antagonistic scenarios that delimit a \textit{robustness window} as a function of dataset size.

\subsection{Learning as Optimal Two-Part Compression}
\label{subsec:two_part}
Consider a dataset of i.i.d. samples $\mathcal{D}_N = \{(x_1, y_1), \dots, (x_N, y_N)\}$ sampled  from an unknown distribution $p^*(x,y)$. We will further assume that the distribution $p^*(x,y)$ factorizes as a mixture $\sum_e p^*(e)\,p^*(x,y \mid e)$ over discrete environments $e \sim p^*(e)$. Each environment-specific distribution $p(x, y \mid e)$ reflects potentially different associations between spurious features in $x$ and the label $y$. Since this paper aims to study the interactions between compression and OOD generalization, we focus on distribution shifts arising from changing $p^*(e)$ at test time.

We adopt the MDL perspective, where learning is equivalent to finding the shortest description of the training dataset $\mathcal{D}_N$. In our supervised setting, the corresponding compression task consists of encoding the labels $y$ given the input $x$. In the two-part version of the MDL principle, we compress the labels sequentially by entropy-coding each $y$ using a candidate conditional probabilistic model $p(y\mid x)$ selected from a hypothesis class $\mathcal{M}$. Because we don't know a priori which model in $\mathcal{M}$ was used to encode the data, the model itself must be encoded as well. The two-part MDL procedure consists then of two steps:
\begin{enumerate}
    \item \textbf{Model Encoding:} We first encode the model $p$ itself. The shortest description length of $p$ is given by its Kolmogorov complexity, denoted $K(p)$. Kolmogorov complexity implies optimal compression, but if some other coding strategy $c$ is used to encode the model, then we can denote the description length of the model under $c$ as $L_c(p)$. We will see in \cref{subsec:desc_length_estimation} how we can compute $L_c(p)$, using a method called prequential coding.
    \item \textbf{Data Encoding:} Using the model $p$, we then encode each label at a cost of $-\log_2p(y\mid x)$ bits, using methods categorized as entropy coding.
\end{enumerate}
Consequently, the total cost to encode the dataset $\mathcal{D}_N$ using a candidate model $p$ is:
\begin{equation}
    \mathcal{J}(p, \mathcal{D}_N) = \underbrace{L_c(p) \vphantom{\sum_{(x,y) \in \mathcal{D}_N}}}_{\text{model cost}} + 
    \underbrace{\sum_{(x,y) \in \mathcal{D}_N} -\log p(y\mid x)}_{\text{data cost given model}}
\end{equation}
To characterize the learning dynamics in a way that is independent of a specific dataset realization, we model neural networks as MDL-optimal compressors in expectation. In this idealized view, we denote by $\mathcal{L}$ the learner that selects the model $p \in \mathcal{M}$ which minimizes the \textit{expected} two-part description length over datasets of size $N$ drawn from the true distribution $p^*$.

By taking the expectation over the data-generating process $\mathcal{D}_N \stackrel{\text{iid}}{\sim} p^*$, the objective function decomposes into interpretable information-theoretic quantities:
\begin{align}
    \mathbb{E}&_{\mathcal{D}_N \sim p^*} \left[ L_\mathcal{L}(p) + \;\sum_{(x,y) \in \mathcal{D}_N}\! -\log p(y\mid x) \right]  \\
    = \; &L_\mathcal{L}(p)  + N \cdot \mathbb{E}_{(x,y) \sim p^*} [-\log p(y\mid x)]    \label{eq:mdl_objective} \\
    = \; &L_\mathcal{L}(p)  + N \cdot \mathbb{E}_{x \sim p^*(x)}\; H(p^*_x,p_x)  \\
    = \; &L_\mathcal{L}(p)  + N \cdot \mathbb{E}_{x \sim p^*(x)}\!\left[ H(p^*_x) + D_{KL}\big(p^*_x \| p_x\big) \right]
\end{align}
where $p_x = p(\cdot\mid x)$ and $p^*_x = p^*(\cdot\mid x)$. $H(p^*_x)$ is the entropy of the labels under the true conditional distribution -- the irreducible epistemic uncertainty in the labels -- and $D_{KL}(p^*_x \| p_x)$ is the average amount of excess bits required to encode a label $y\sim p^*_x$ using $p_x$ instead of the true conditional $p_x^*$, measuring the misalignment between the hypothesis and the true conditional distribution.

Since $H(p^*)$ is constant with respect to $p$, the learner effectively solves:
\begin{equation}
    \hat{p}_N = \underset{p \in \mathcal{M}}{\arg\min} \bigg[ \underbrace{L_\mathcal{L}(p) \vphantom{\mathbb{E}_{x \sim p^*(x)} D_{KL}(p^*_x \| p_x)}}_{\text{model cost}} + N \cdot \underbrace{\mathbb{E}_{x \sim p^*(x)} D_{KL}(p^*_x \| p_x)}_{\text{excess data cost}}\bigg]
    \label{eq:final_objective}
\end{equation}

\begin{figure*}[htbp]
    \centering
    \includegraphics[width=\textwidth]{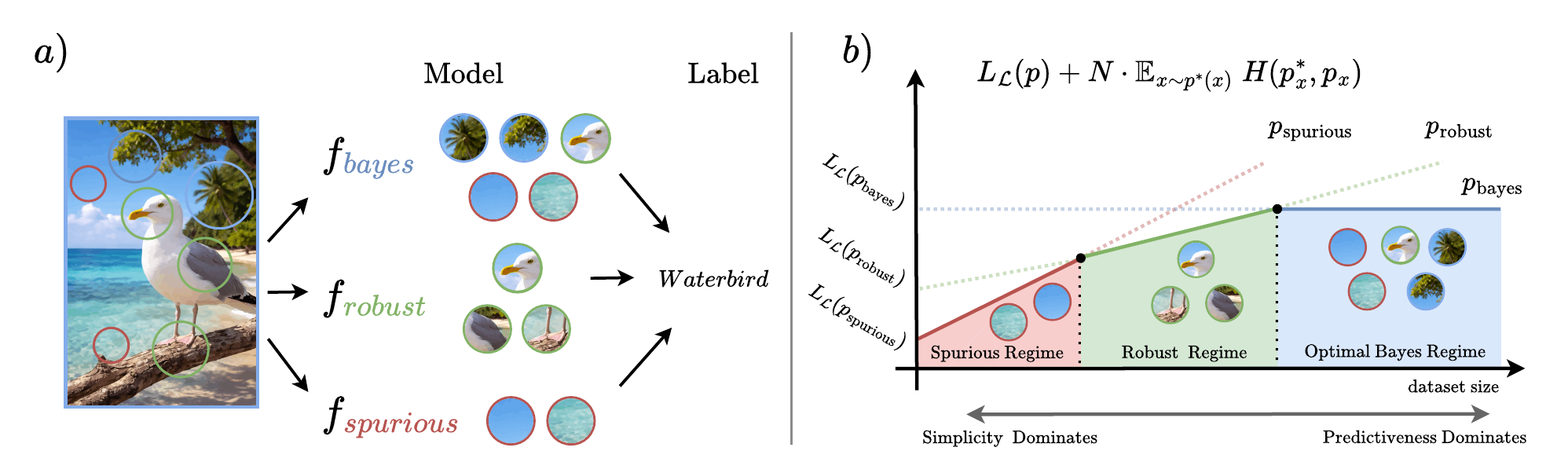}
    \caption{\textbf{(a)} In a supervised setting, some features causally relate to the label (e.g., phenotypical characteristics), while others are spurious (e.g., background of the image). A Bayesian solution leverages all available features.
    \textbf{(b)} Total expected compression cost as a function of training dataset size $N$. Each line represents the two-part description length of a candidate model. The MDL-optimal learner selects the model achieving the lowest total cost at each $N$, inducing transitions between qualitatively different solutions as the amount of data increases.}
    \label{fig:compression}
\end{figure*}

This formulation explicitly quantifies the trade-off between model complexity\footnote{$L_\mathcal{L}$ depends on $\mathcal{L}$, since the learning algorithm can contain arbitrary prior knowledge about the task in its primitives that impact the description length of $p$.} and average goodness-of-fit. Because this objective is a linear function of $N$, the optimal hypothesis $\hat{p}_N$ necessarily shifts qualitatively as the volume of training data grows. We provide a visual illustration of this phenomenon in \cref{fig:compression}. This compression-centric view provides a \textbf{principled account of how data scarcity} (through $N$) \textbf{and the characteristics of different candidate predictors $p \in \mathcal{M}$} (through $L_\mathcal{L}(p)$ and $D_{KL}(p^*_x \| p_x)$) \textbf{govern the preference of an MDL-optimal learner}. 

\subsection{Analysis of Learning Regimes} 
\label{subsec:analysis}

Recalling that the idealized learner $\mathcal{L}$ chooses the model with the shortest total codelength, the decomposition above shows how the learner's preference for a solution $p$ shifts as the amount of training data $N$ increases (see \cref{fig:compression}): the total expected description length is the sum of a \textbf{fixed bit-cost} $L_\mathcal{L}(p)$ and a \textbf{variable bit-cost} that grows linearly with the amount of data $N$. Depending on the data regime, different components of the total description length dominate:
\begin{itemize}
    \item \textbf{Low-Data Regime:} The fixed cost $L_\mathcal{L}(p)$ dominates. The learner prioritizes models with short description length (i.e., \textit{simpler} models) even if they only incompletely capture the structure in the data. This can result in overfitting -- memorization of the few training datapoints -- or reliance on simple spurious features.
    \item \textbf{High-Data Regime:} The variable linear cost of encoding labels $N\cdot  \mathbb{E}_{x\sim p^*(x)} D_{KL}(p^*_x \| p_x)  $ dominates. The idealized learner $\mathcal{L}$ is driven to minimize the conditional Kullback-Leibler (KL) divergence for all $x$ in the support of $p^*(x)$, selecting the most predictive model $p_x^*$ (or its closest approximation in $\mathcal{M}$), regardless of its complexity. Despite learning the true generative process in the limit of infinite data, this can also result in reliance on complex environment-specific features that may not be robust under distribution shifts (e.g., observations under a novel environment at test time).
\end{itemize}

The analysis above shows how the MDL-optimal compressor's preference is not fixed: it depends on the dataset size $N$. As $N$ grows, preference shifts from simple, rudimentary hypotheses toward more predictive but complex ones. Assuming modern learning algorithms behave as MDL-optimal compressors \citep{goldblum2023no, wilson2025deep}, this observation has direct implications for practitioners because in many real-world tasks inputs carry multiple predictive features -- some causal and robust to distribution shift, others spurious and environment-dependent -- and the learner implicitly chooses among them. Selecting features that generalize beyond the training distribution is notoriously difficult, precisely because spurious features can be both simpler and sufficiently predictive of the training data. Our framework suggests that this selection is governed by a compression trade-off: the feature the learner relies on at any given $N$ is the one whose corresponding model minimizes the total description length. This perspective offers a principled account of \textit{what drives feature selection across data regimes}, and can be used to predict when a learner will favor robust features over spurious shortcuts, or vice versa. In \cref{sec:experiments,sec:results}, we empirically test these predictions on a controlled benchmark.

\subsection{Implications for Robust Learning}
\label{subsec:robust}
Now that we have introduced the objective function of our idealized learner $\mathcal{L}$ and discussed how this formulation accounts for feature selection across data regimes, we next discuss its implications for robust learning.

Recall that inputs $x$ are generated from a mixture of environments $e \in \mathcal{E}$, each inducing different correlations between features and labels. In this setting, $x$ simultaneously carries (i) causal features (e.g., feather, spout, claw) that are invariant across environments, (ii) simple spurious features (e.g., ocean texture) whose correlation with labels varies by environment, and (iii) environment-specific signatures from which $e$ itself can potentially be inferred. Each type of feature gives rise to a qualitatively different predictive model, and the compression framework in \cref{subsec:two_part} prescribes which one the idealized learner selects at a given $N$. We now discuss the compression trade-off for three archetypal models that span this feature spectrum:

\begin{itemize}
    \item \textbf{Robust Model ($p_{\text{robust}}$):} Uses only invariant, causal features  (e.g., animal attributes). Moderate complexity, but generalizes across environments.
    \item \textbf{Bayes-Optimal Model ($p_{\text{bayes}}$):} Integrates all available information and predicts via the posterior predictive distribution:
    \[
        p_{\text{bayes}}(y\mid x) \,=\, \sum_{e \in \mathcal{E}} p(y\mid x,e)\, p(e\mid x)
    \]
    This achieves minimal empirical error, but can fail under distribution shift.
    \item \textbf{Spurious Model ($p_{\text{spur}}$):} Uses simple, non-robust features (e.g., background textures) and marginal environment probability $p(e)$. Low complexity, but fails under distribution shift. The spurious model is a degenerate Bayesian model: when the environment cannot be inferred from $x$, the predictor defaults to using the marginal $ p_{\text{spur}}(y\mid x) \,=\, \sum_{e \in \mathcal{E}} p(y\mid x,e)\, p(e)$.
\end{itemize}

We consider two opposing scenarios relevant to practitioners. In each, the causal solution appears either as a simple solution with limited predictiveness, or as the most predictive but also more complex solution. For each scenario, we estimate the total description length of competing solutions and test whether, as we sweep across data regimes, the predicted solution of our idealized learner -- i.e.\ the one that yields the lowest total description length -- matches the neural network's learned solution, as reflected by shifts in feature reliance and generalization performance (see \cref{fig:compression}).

\paragraph{Scenario A: Spurious vs. Robust solution}
In this scenario, the robust model $p_{\text{robust}}$ is the most predictive candidate for modeling the data, but it is also more complex than available spurious solutions ($L_\mathcal{L}(p_{\text{robust}}) > L_\mathcal{L}(p_{\text{spur}})$ but $D_{KL}(p^* \| p_{\text{robust}}) < D_{KL}(p^* \| p_{\text{spur}})$). This is the classic ``shortcut learning'' case where a simple feature (e.g., a background texture), usually benefiting from strong environment imbalance, arises as a ``good-enough'' predictor within the amount of training data.

\begin{itemize}
    \item \textbf{Behavior:} At low $N$, the idealized learner favors $p_{\text{spur}}$ because its low description cost $L_\mathcal{L}(p_{\text{spur}})$ outweighs its higher divergence $D_{KL}(p^* \| p_{\text{spur}})$. The transition to the robust model only occurs once $N$ is large enough to \textit{pay} for the complexity overhead of $p_{\text{robust}}$. Said differently, increasing $N$ makes the selection of the robust model increasingly appealing.
    \item \textbf{Transition Dynamics:} The amount of data required to reach the robust regime \textit{increases} with the complexity gap $L_\mathcal{L}(p_{\text{robust}}) - L_\mathcal{L}(p_{\text{spur}})$, and \textit{decreases} as the predictive advantage of robust features over spurious ones grows, which can be achieved for instance by increasing the diversity of training environments.
\end{itemize}

\paragraph{Scenario B: Robust vs. Bayes-Optimal solution}
In this scenario, the comparative advantage of the robust model is reversed: it is \textit{not} necessarily the most predictive model, but stands as a simpler alternative to the more complex Bayes-optimal model $p_{\text{bayes}}$ that uses all available latent features in the input $x$ to infer the environment $e$ and achieve minimal empirical risk. 

\begin{itemize}
    \item \textbf{Behavior:} If the Bayes-optimal solution is significantly more complex -- for instance, because environment inference requires extracting complex background features -- then at low $N$, the marginal predictive gain of $p_{\text{bayes}}$ does not yet justify its higher description cost $L_\mathcal{L}(p_{\text{bayes}})$. However, as $N \to \infty$, the learner will inevitably transition to the best predictive model $p_{\text{bayes}}$, which may fail under OOD conditions such as unseen environments.
    \item \textbf{Transition Dynamics:} The amount of data required to transition \textit{away} from the robust regime to the Bayes-optimal one \textit{increases} with the complexity gap $L_\mathcal{L}(p_{\text{bayes}}) - L_\mathcal{L}(p_{\text{robust}})$, and \textit{decreases} with the predictive advantage of environmental cues $D_{KL}(p^* \| p_{\text{robust}}) - D_{KL}(p^* \| p_{\text{bayes}})$.
\end{itemize}

Taken together, these two scenarios illustrate the \textbf{nuanced role of simplicity bias}, where the interplay between model complexity and predictive power determines the appropriate data regime for robustness. Specifically, Scenario A defines a lower bound $N_{\text{min}}$ on the data required to overcome simplicity bias and rule out spurious shortcuts, while Scenario B defines an upper bound $N_{\text{max}}$ beyond which the learner will sacrifice robustness for the superior predictive power of a more complex, environment-dependent solution. Ultimately, this information-theoretic perspective underscores that simplicity bias is a double-edged sword: it can either hinder generalization by favoring shortcuts or promote it by ruling out overly complex environment-dependent hypotheses. 

\section{Experimental Setting}
\label{sec:experiments} 
In this section, we now aim to empirically test the hypothesis that neural networks behave as MDL-optimal compressors as posited in \cref{sec:theory}. Our goal is to test whether the data-regime transitions predicted by our theory, i.e.\ the crossover points where one hypothesis begins to yield a shorter total description length than another (see \cref{fig:compression}b), line up with empirical shifts in feature reliance and generalization performance of a trained neural network. To this end, we first (i) introduce a semi-synthetic visual benchmark where we can precisely control aspects of the data-generating process, such as feature complexity and predictiveness. We then (ii) show how to estimate the total compression cost \eqref{eq:mdl_objective} of a candidate model $p$ in practice. Finally, we (iii) present evaluation metrics for feature reliance used to confirm whether the learned model $p_N$ relies on specific features as predicted by our theory.
\subsection{Benchmark Design}
\label{subsec:design}

To validate our theoretical framework, we design a semi-synthetic visual task derived from Colored MNIST \citep{arjovsky2020invariantriskminimization} (see \cref{fig:task_samples} for examples): The task is to predict whether a handwritten digit is smaller or greater than $5$ where each sample is drawn from a mixture of discrete environments. Every input simultaneously carries three types of features: (i) the \textbf{digit shape}, the causal determinant of the label; (ii) an  environment-dependent \textbf{color} applied to the digit that spuriously correlates with labels; and (iii) a \textbf{binary watermark} drawn from an environment-specific bank of $K$ random patterns embedded in the rightmost pixel column of the image. The predictiveness of each feature is controlled by independent noise parameters, and the bank size $K$ directly controls the complexity of the watermark feature. See \cref{appendix:task_details} for the full data-generating procedure and the role of each parameter. By varying these knobs, we explore the full span of scenarios identified in \cref{sec:theory}:

\begin{itemize}
    \item \textbf{Scenario A: Spurious vs. Robust solution} The spurious color signal acts as an easily exploitable shortcut; it is simpler to encode than the robust digit features ($L_\mathcal{L}(p_{\text{spur}}) < L_\mathcal{L}(p_{\text{robust}})$), despite usually being less predictive of the label. This specifically mirrors cases where the environment is not directly inferrable from the input, but a skewed marginal environment distribution $p(e)$ allows the learner to achieve decent predictive performance through a trivial feature (e.g., $\text{water} \Leftrightarrow \text{water-bird}$). This allows us to study simplicity bias in its most common form: the preference for a trivial but non-robust signal over a more sophisticated causal mechanism.
    \item \textbf{Scenario B: Robust vs. Bayes-optimal solution} Watermarks (drawn from environment-specific banks) act as highly predictive features, but complex to learn since exploiting them requires memorizing an arbitrary number of distinct watermark patterns. Here, our theory predicts a protective effect: the learner is prevented from exploiting these signals until $N$ is large enough to justify the high encoding cost of $L_\mathcal{L}(p_{\text{bayes}})$.
\end{itemize}

\subsection{Estimating the Total Compression Cost}
\label{subsec:desc_length_estimation}
With the benchmark defined, we now describe how the MDL predictions are computed. The two-part MDL objective \eqref{eq:mdl_objective} decomposes into a fixed-cost and a variable cost, and one of the main challenges is to find tractable methods to estimate these two components for any $p$. We detail the estimation procedure for each component below.

\paragraph{Fixed cost: $L_\mathcal{L}(p)$} The candidate predictors $p_{\text{spur}}$, $p_{\text{robust}}$, and $p_{\text{bayes}}$ introduced in \cref{subsec:robust} are abstract objects defined by the features they exploit. To obtain a concrete instantiation whose description length can be computed, we operationalize each candidate $p_\texttt{feature}$ as a neural network $p_N$ trained on a custom dataset $\mathcal{D}_N$, where all variables except \texttt{feature} are randomly shuffled, which guarantees that only \texttt{feature} gets extracted in the process. When a model is obtained through a learning procedure, its description length $L_\mathcal{L}(p_N)$ can be tightly estimated via \textit{prequential coding} \citep{blier2018description}. In this view, simplicity relates to \textit{ease of learning}: a simple model is one that can be learned quickly -- with just a few samples. Its description length corresponds to the cumulative excess log-likelihood on unseen data incurred before the model has enough training data to converge, which requires $N$ to be large enough in practice. We provide the full derivation in \cref{appendix:prequential_details}. 

\paragraph{Variable cost: $ N \cdot \mathbb{E}_{(x,y) \sim p^*} [-\log p(y\mid x)]$} The variable cost corresponds to the expected number of bits required to encode labels sampled from the true distribution $p^*$ using the model $p$, it corresponds to the remaining intrinsic randomness of the data that cannot be modeled by $p$. This quantity is computed as the empirical cross-entropy loss of the candidate model $p$ evaluated on a held-out dataset $\mathcal{D}_{\text{test}}$ sampled from $p^*$:
\[
    \mathbb{E}_{(x,y) \sim p^*} [-\log p(y\mid x)] \,\approx\, \frac{1}{|\mathcal{D}_{\text{test}}|} \sum_{(x,y) \in \mathcal{D}_{\text{test}}} -\log p(y\mid x)
\]

Finally, for each candidate model $p$, it is also possible to estimate the fixed and variable costs of \textit{intermediate models} $\{p_1, \cdots ,p_N\}$ that have not yet converged. We provide a detailed explanation in \cref{appendix:intermediate_models}. Intuitively, those intermediate models have absorbed fewer bits of structure from the data, yielding lower complexity but higher compression rate; together they define an \emph{envelope} of compression lines.

\subsection{Evaluation Metrics}
In order to detect the phase transition in the learning procedure, we develop a set of metrics designed to quantify the model's reliance on specific input features through evaluation on custom OOD datasets.
\begin{enumerate}
    \item \textbf{Accuracy on held-out data}: We evaluate the model's performance across different data distributions. \texttt{Training} and \texttt{Validation} respectively track memorization and ID generalization ability of the trained model. We introduce \texttt{Digit}, \texttt{Color} and \texttt{Watermark} test sets, where the label correlates exclusively with a single \texttt{feature}, measuring the model's reliance on that specific \texttt{feature}. By breaking the correlation between the label and complementary features, these tests measure OOD generalization under targeted covariate shifts.
    \item \textbf{Permutation feature importance}: We assess model reliance on each feature by randomly shuffling the values of a single feature of interest (e.g., digit, color, or watermark) across the test set, breaking its correlation with the label while preserving its marginal distribution. The \emph{accuracy gap}, defined as the drop in accuracy between the original and permuted datasets, quantifies how much the model depends on that feature for prediction.
\end{enumerate}

\section{Results}
\label{sec:results}
In \cref{subsec:qualitative}, we present a side-by-side comparison of the MDL-optimal compression view and the empirical behavior of trained neural networks on Scenarios A and B, directly testing the core claim of the paper: that the feature transitions predicted by the compression envelope coincide with empirical shifts in feature reliance. In \cref{subsec:quantitative}, we provide quantitative evidence that varying feature predictiveness and complexity impacts feature selection as predicted, supporting our central hypothesis.
\subsection{Dissecting a Single Experiment}
\label{subsec:qualitative} 
We begin by examining Scenarios A and B, where the same task is modeled both as a problem of optimal compression and as a problem of learning. For each scenario, we compare the preferences of an MDL-optimal compressor side-by-side with those of a trained neural network. The results are reported in \cref{fig:qualitative_study}. We provide a detailed description of the experimental setup in \cref{appendix:experiment_details}.
\paragraph{Compression view}
We first estimate the total description length for a set of candidate predictors and their intermediate interpolations, each corresponding to a distinct \texttt{feature} type. The fixed cost $L_{\mathcal{L}}(p_\texttt{feature})$ is estimated with prequential code length on a custom dataset as explained in \cref{subsec:desc_length_estimation}, while the variable cost $N \cdot \mathbb{E}_{(x,y) \sim p^*}[-\log p(y\mid x)]$ is estimated via cross-entropy on the original test dataset. Plotted on a log-log scale (\cref{fig:qualitative_study}a), the resulting \emph{compression lines} reveal for any given dataset size $N$ which solution locally yields the shortest description length and show the predicted transition point (vertical marker) where the MDL envelope shifts from one dominant feature type to another. In this envelope, each line corresponds to a model that extracts the corresponding feature with increasing accuracy: for example, there is a range of models that extract digit features, going from lossily extracting some simple geometric primitives to learning the full subtlety of handwriting.

\paragraph{Learning view}
In parallel, we train randomly initialized neural networks for different training dataset sizes $N$. We then evaluate them alongside the solutions selected by the MDL-optimal compressor on the set of metrics described in \cref{subsec:design}. These metrics reveal which feature those models actually rely on at each data regime (\cref{fig:qualitative_study}b--c). We provide additional details about model class, optimizer and training in \cref{appendix:training_hyperparameters}.
\paragraph{Alignment}
Comparing both views, we highlight two findings. First, the \emph{hard transition} predicted by the compression envelope -- i.e.\ the dataset sizes at which the MDL-optimal solution switches from one dominant feature to another -- coincide precisely with the empirical transition in feature reliance. Second, while the OOD performance of the theoretical solution matches that of the neural network on the dominant feature, our current framework evaluates candidate models that exploit strictly disjoint subsets of features. Consequently, the idealized MDL solution exhibits discrete transitions in feature reliance (\cref{fig:qualitative_study}b), entirely discarding a previously favored feature. In practice, however, neural networks shift continuously between features. In the data regime surrounding a transition, there may exist a hybrid model that partially exploits the complex feature while still leveraging residual signal from the simpler one, leading to a gradual crossover around the predicted transition point.

\begin{figure}[h]
    \centering
    \includegraphics[width=0.48\textwidth]{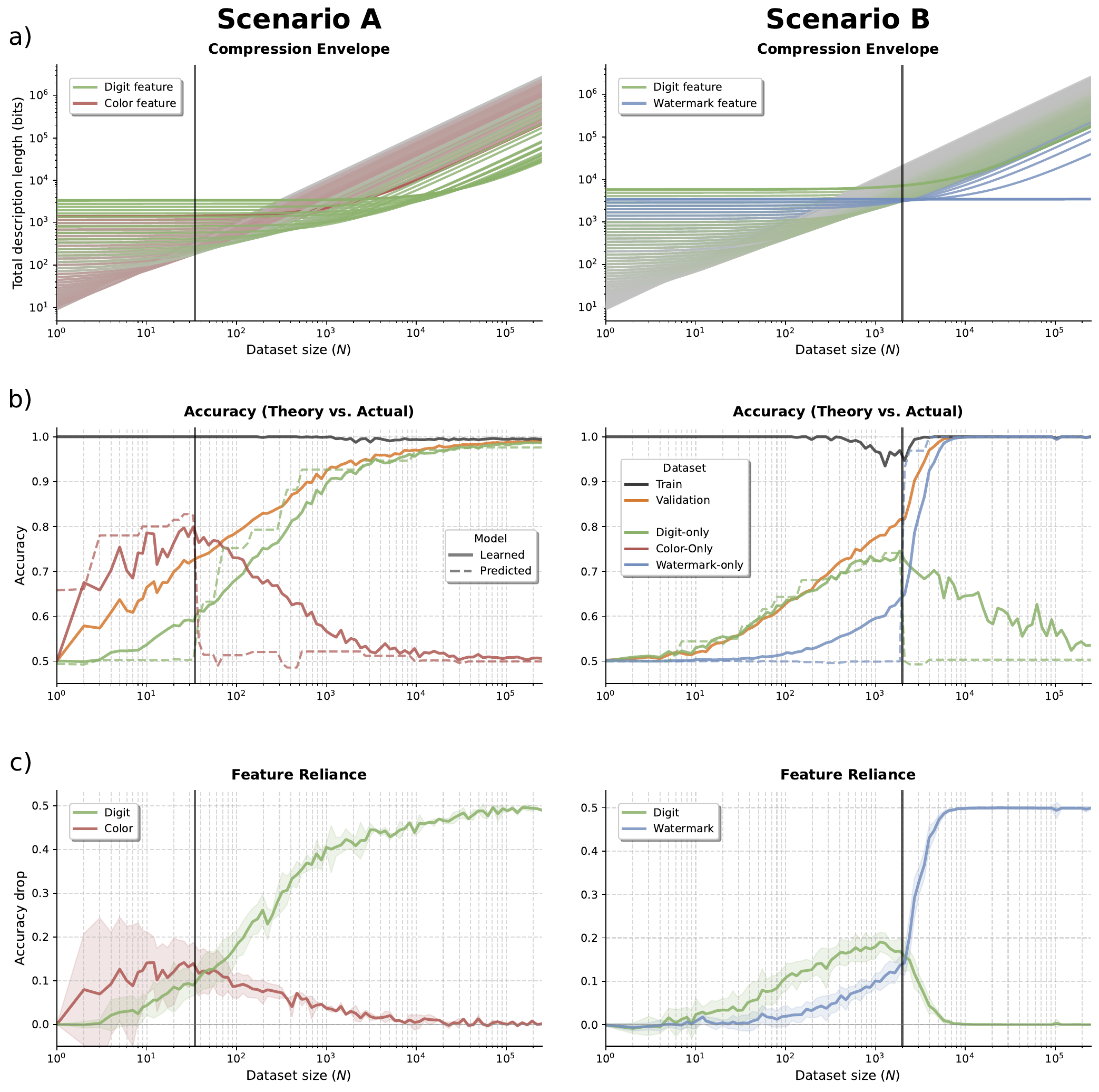}
    \caption{Comparison of the compression view and the learning view for Scenario A and B. \textbf{(a)} Compression envelopes on a log-log scale: each line shows the total two-part description length of a candidate predictor as a function of dataset size $N$. \textbf{(b)} Accuracy on held-out datasets as a function of $N$: solid lines are empirical measurements, dashed lines are reported metrics of the MDL-optimal solution. \textbf{(c)} Permutation feature importance (accuracy drop after shuffling each feature): a higher drop indicates stronger reliance on that feature.}
    \label{fig:qualitative_study}
\end{figure}

\subsection{Varying task characteristics}
\label{subsec:quantitative}
\begin{figure*}[htbp]
    \centering
    \begin{subfigure}[t]{0.49\textwidth}
        \centering
        \includegraphics[width=\textwidth]{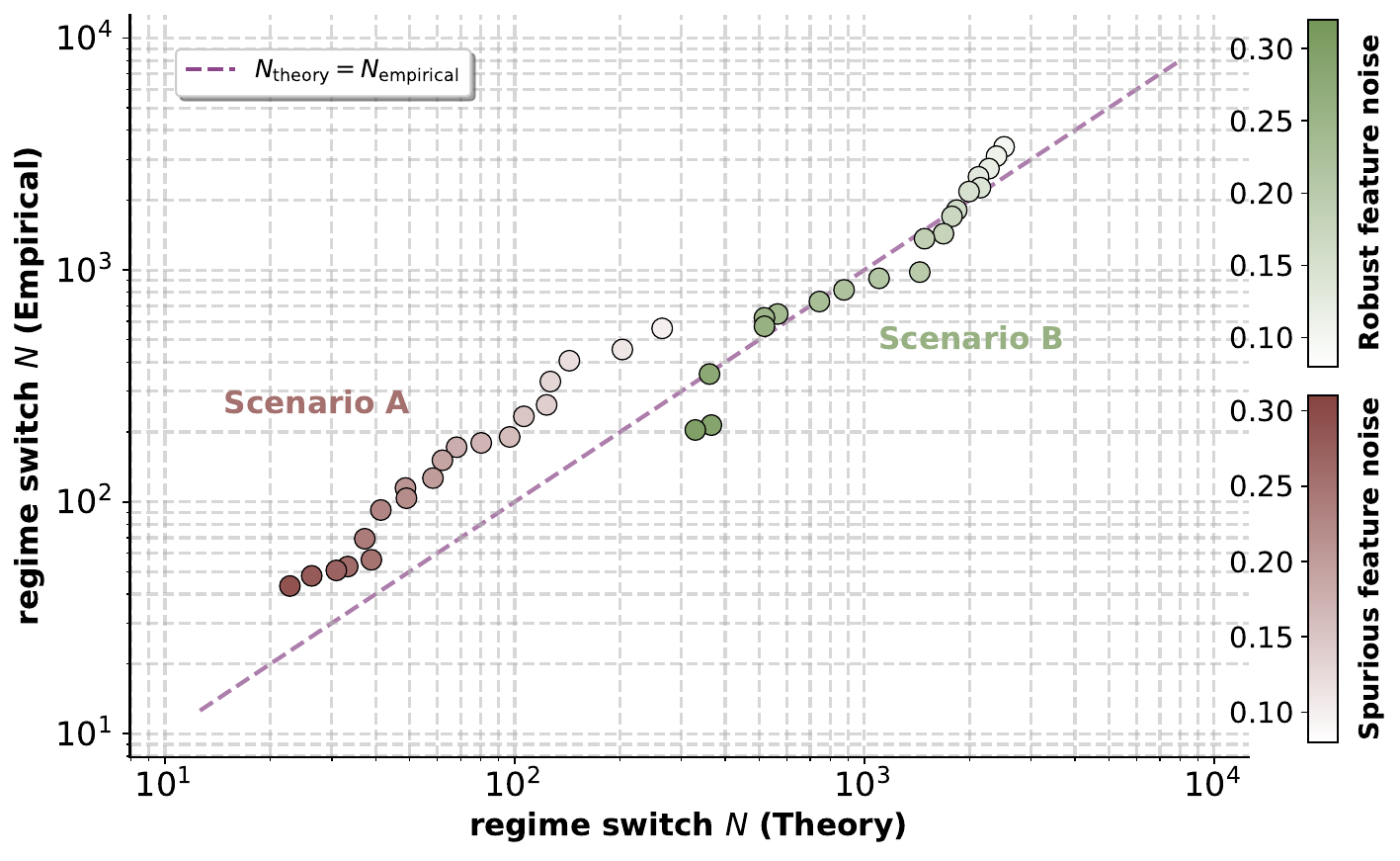}
        \caption{Effect of feature predictiveness}
        \label{fig:scatter_slope}
    \end{subfigure}
    \hfill
    \begin{subfigure}[t]{0.49\textwidth}
        \centering
        \includegraphics[width=\textwidth]{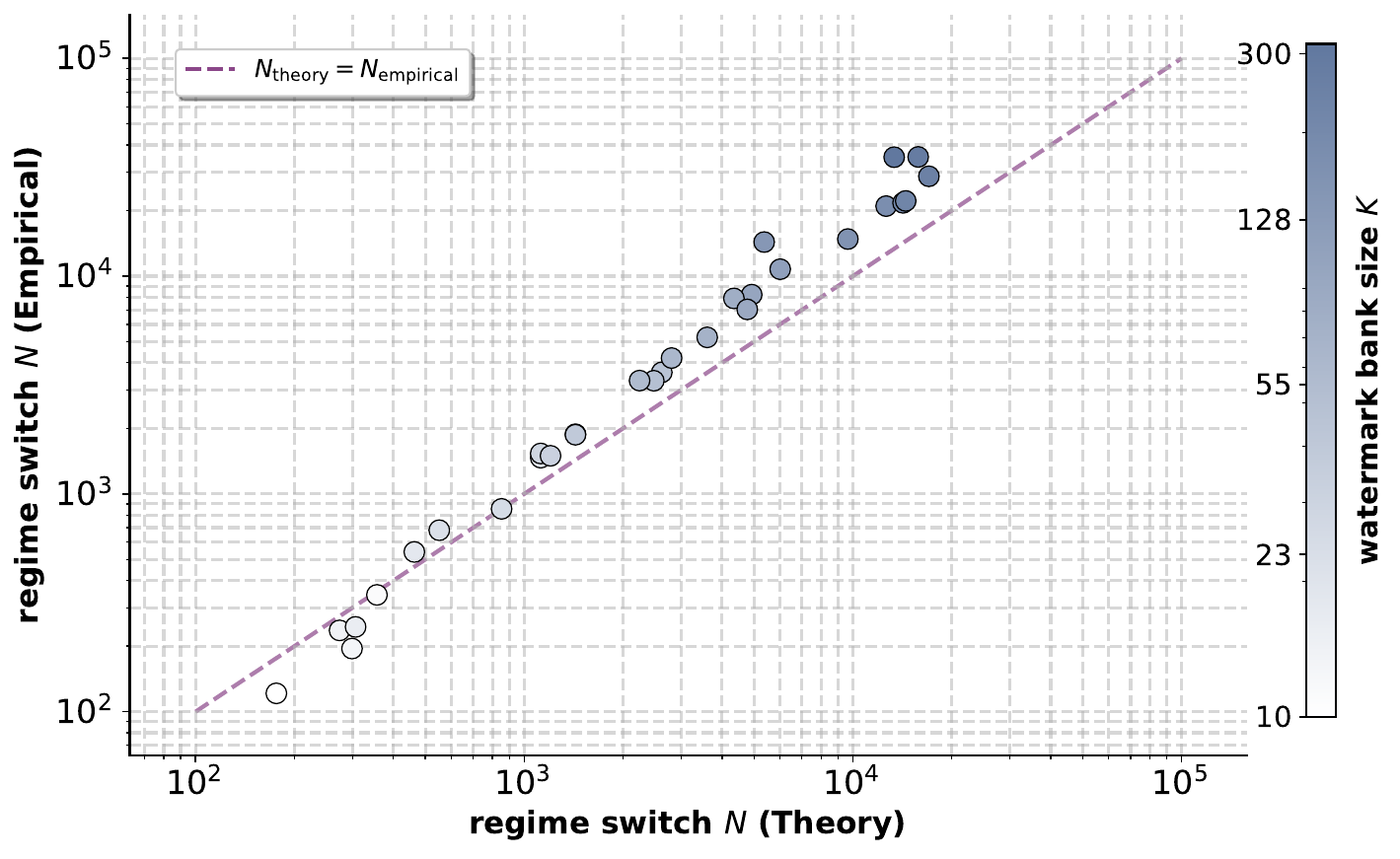}
        \caption{Effect of feature complexity}
        \label{fig:scatter_complexity}
    \end{subfigure}
    \caption{Empirical vs.\ theoretical transition point $N$ for varying task characteristics. Each point is one experimental configuration; the dashed line is the identity $N_{\text{theory}} = N_{\text{empirical}}$. \textbf{(a)} Reducing the predictiveness of a feature increases its compression rate disadvantage, causing the transition to occur earlier. \textbf{(b)} Increasing the complexity of a feature inflates its description cost, causing the transition to occur earlier. In both cases, the proximity to the diagonal indicates that our MDL-based theory accurately predicts the observed feature switch as measured by our proxy.}
    \label{fig:scatter_quantitative}
\end{figure*}

In this second part, we now study the effect of varying key task characteristics such as feature complexity and correlation strength on the feature selection behavior of neural networks, and compare the observations with the predictions derived from our theory. Results are summarized in \cref{fig:scatter_quantitative}. For each configuration, we extract two scalar statistics:
\begin{itemize}
    \item \textbf{Theoretical transition $N_{\text{theory}}$}: the dataset size at which the MDL compression envelope switches between different feature types, i.e.\ the point where the two-part codelength of one feature-based predictor falls below that of the competing one. 
    \item \textbf{Empirical transition $N_{\text{empirical}}$}: the dataset size at which the trained neural network's dominant feature reliance switches, detected as the crossover point where the accuracy gap of one feature overtakes the other.
\end{itemize}
\paragraph{Effect of feature predictiveness:}
In both scenarios, reducing the predictiveness of a feature lowers its compression rate advantage over the competing solution, causing the transition to occur earlier. In Scenario~A, we vary the noise on the spurious color/label correlation: a noisier spurious feature increases its per-sample bit-cost, so the idealized learner abandons it sooner and $N_{\text{theory}}$ decreases (\cref{fig:scatter_slope}). Symmetrically, in Scenario~B, as the predictiveness of the robust model decreases, the gap in compression rate between the robust and the Bayes-optimal model increases. This widening gap rapidly eclipses the robust model's initial cost advantage, causing the transition point $N_{\text{theory}}$ to decrease (\cref{fig:scatter_slope}).

\paragraph{Effect of feature complexity:}
Finally, in Scenario~B we vary the number of distinct watermark patterns the model must learn to exploit the complex watermark/label correlation, which necessarily inflates $L_{\mathcal{L}}(p_{\text{bayes}})$. Concretely, larger banks should delay the transition away from the robust regime: $N_{\text{theory}}$ increases (\cref{fig:scatter_complexity}).

Results show that $N_{\text{theory}}$ and $N_{\text{empirical}}$ correlate well across all configurations, with a Pearson correlation of $0.976$, especially considering the noise in the estimation of the cross-over points and model complexity $L_{\mathcal{L}}(p)$. This supports that our MDL-based theory captures the main factors driving feature selection across data regimes.

\section{Discussion}
 
\paragraph{Simplicity and generalization} As we noticed in the experiment Section, simplicity bias does not necessarily promote nor prevent generalization. Paradoxically, in low-data regimes, it can encourage overfitting and reliance on spurious shortcuts as the `most compact' explanation for the finite set of samples\footnote{although the No Free Lunch theorem suggests that no learning algorithm likely provides generalization guarantees in such low-data regimes} (\cref{fig:qualitative_study}). Importantly, this is not a failure of the learner: from the MDL perspective, memorization is the \emph{rational} solution when the available data are too few to justify the description cost of any structured model. As the number of samples grows, however, the cumulative extra-cost in bits of memorizing samples or using a suboptimal model $p$ becomes increasingly unbearable, eventually forcing the learner to switch to more predictive competing models, regardless of their inherent complexity. 

\paragraph{A complexity-based regularization} An important implication arises when the robust model is not the most predictive candidate -- a common scenario where environment-specific features are highly predictive within the training distribution. In such cases, a more complex Bayes-optimal model incorporating these non-robust features would achieve a lower compression rate asymptotically. However, our theory predicts that a simplicity-seeking learner will not select this complex environment-dependent model while the dataset size $N$ remains below a theoretical threshold such that the marginal predictive gain of the Bayes-optimal solution has not yet offset its higher description cost. This is precisely what we observe in Scenario~B, where reliance on the robust digit feature (\cref{fig:qualitative_study}c, green line) peaks in the intermediate data regime before the model abandons it in favor of the more predictive but more complex Bayesian solution. Paradoxically, \textbf{limiting the amount of training data can act as a form of complexity-based regularization}: a constrained data budget keeps the learner in the regime where simpler robust mechanisms are favored, because the more predictive alternatives remain unaffordable from a compression standpoint. 

Finally, our framework also suggests a compression-theoretic rationale for pretraining: through unsupervised exposure to diverse environments, a pretrained model has already absorbed, for free, bits of structure about the data into its weights, which lower the operational description cost $L_\mathcal{L_\text{pretrain}}(p_{\text{robust}})$ before fine-tuning begins, making complex solutions accessible at smaller dataset sizes \citep{hendrycks2019using}.


\section{Related Work}
\label{sec:related_work}

This paper adopts an information-theoretic perspective on machine learning, grounded in the established equivalence between statistical learning and lossless compression \citep{shannon1948mathematical, rissanen1978modeling,chaitin2002intelligibility}: to \textit{comprehend} the underlying structure of a dataset is identical to \textit{compressing} that dataset into a more efficient representation. This duality has birthed an information-theoretic paradigm for model selection with applications to robust machine learning: the Minimum Description Length (MDL) principle \citep{grunwald2007minimum}.

This paper relates most closely to a line of papers that empirically analyze the tendency of NNs to prefer simple features for prediction, referring to this phenomenon as simplicity bias \citep{rahaman2019spectral,de2019random}, and their implications for ID and OOD generalization. The simplicity bias of neural networks presents a fundamental duality: while it is an instrumental mechanism for ID generalization, it is simultaneously the root cause of systemic failure of OOD generalization. On the one hand, SB relates to robust ID performance by favoring the discovery of underlying patterns and preventing memorization or the learning of redundant features \citep{arpit2017closer, valle2018deep, pezeshki2021gradient, mingard2025deep}, aligning with Occam's razor. On the other hand, this same preference for simplicity often compels models to rely on spurious correlations and low-level shortcuts \citep{geirhos2018imagenet, geirhos2020shortcut, shah2020pitfalls, pezeshki2021gradient}. Because these non-robust rules are typically ``simpler'' to optimize, NNs remain vulnerable to adversarial samples that exploit spurious covariate shifts \citep{teney2022evading, vasudeva2023mitigating}.

\section{Limitations \& Future Work}
\label{sec:limitations}

Our framework currently considers candidate models that exploit strictly disjoint subsets of features, some of which are perfectly predictive. While this idealization enables a clean theoretical analysis, real-world tasks typically involve multiple \emph{partially} predictive features that contribute to label prediction, interact with one another, and may not be neatly separable. In such settings, the compression trade-off is richer: the MDL-optimal learner may blend several features simultaneously rather than transition sharply between them, and the notion of a single ``dominant'' feature may not hold. Extending the framework to predictors that handle multiple partially predictive features would bring the theory closer to realistic settings and enable the study of more nuanced generalization failure modes.

\begin{acknowledgements}
The authors would like to thank Avery Hee-Woon Ryoo for their useful comments. 
TM acknowledges support from the Fonds de recherche du Qu\'{e}bec (FRQNT 0000364322, \url{https://doi.org/10.69777/0000364322}). EE acknowledges support from Vanier Canada Graduate Scholarship \#492702. SM acknowledges the support of PhD Excellence Scholarship from UNIQUE. DS acknowledges support from NSERC Discovery Grant RGPIN-2023-04869, and a Canada-CIFAR AI Chair. GL acknowledges support from NSERC Discovery Grant RGPIN-2018-04821, the Canada Research Chair in Neural Computations and Interfacing, and a Canada-CIFAR AI Chair. This research was enabled in part by computing resources and support provided by Mila -- Quebec AI Institute.
\end{acknowledgements}







\bibliography{references}

\clearpage
\onecolumn

\appendix

    \counterwithin{figure}{section} 
    \counterwithin{table}{section}  

\section{Prequential coding details}
\label{appendix:prequential_details}

Prequential coding is a category of universal coding schemes that builds on the idea of sequentially using past data to encode future data as it arrives. In this scheme, assuming the prior knowledge of a learning algorithm $\mathcal{L}$ (architecture, optimizer, initialization scheme, etc.), we recursively encode a label $y_i$ using entropy coding derived from a parametric conditional model $p_i(\cdot \mid x_i)$ fitted on the already transmitted labeled data $\mathcal{D}_{1:i-1}$. The total cost of encoding data with prequential coding is given by:
\[
    L_{\text{preq}}(\mathcal{D}_N;\mathcal{L}) \,=\, \sum_{i=1}^N -\log p_i(y_i\mid x_i)
\]
With $p_{i} \leftarrow \mathcal{L}(\mathcal{D}_{1:i-1})$ and $p_1$ an initial predictor defined by $\mathcal{L}$. At the end of the procedure, both the sender and the receiver share the knowledge of $\mathcal{D}_{1:N-1}$. A final model $p_N$ can then be fitted by the receiver using $\mathcal{L}$ at \textit{zero additional cost}. In this perspective, prequential coding is a program that jointly encodes the data $\mathcal{D}_{1:N-1}$ and the final predictive model $p_N$. We then estimate the description length of $p_N$ alone by rewriting the prequential codelength as follows:
\begin{align*}
        L_{\text{preq}}(\mathcal{D}_N;\mathcal{L}) \,&=\, \underbrace{\sum_{i=1}^N -\Big( \log p_i(y_i\mid x_i) - \log p_N(y_i\mid x_i) \Big)}_{\text{Excess loss}} \quad +  \quad \underbrace{\sum_{i=1}^N -\log p_N(y_i\mid x_i)}_{\text{Asymptotic Loss}}
\end{align*}
Intuitively, the cumulative excess loss incurred before the predictor has converged corresponds to bits spent on describing the final model $p_N$ through $L_\mathcal{L}$. In this regard, a simple model is a model that is learned \textit{quickly}, i.e., that requires fewer samples to be learned. Visually, the description length of the final model is the ``area'' under the curve above the asymptotic loss:
\[
    L_\mathcal{L}(p_N) \,\approx\, \sum_{i=1}^N \Big( -\log p_i(y_i\mid x_i) + \log p_N(y_i\mid x_i) \Big)
\]

\paragraph{Block-wise approximation.}
Computing the exact prequential codelength would require retraining a model from scratch for every new sample. To make this tractable with neural networks, we use a block-wise approximation. The idea is to only retrain the model at a sparse set of dataset sizes and use the resulting predictor to encode the following block of samples.

Let $1 = t_0 < t_1 < \cdots < t_S = N$ be a sequence of block boundaries. In practice we use exponentially increasing block sizes to minimize the estimation error since training loss usually correlates log-linearly with the amount of training data. At each boundary $t_s$, we train a randomly initialized model on the already transmitted data $\mathcal{D}_{1:t_s-1}$. We then use this model to encode the next block $\mathcal{D}_{t_s:t_{s+1}-1}$.

Formally, let $p^{(s)}$ be the predictor trained on $\mathcal{D}_{1:t_s-1}$ for $s \geq 1$ and $p^{(0)}$ the untrained predictor from $\mathcal{L}$. The block-wise prequential description length is:
\begin{equation}
    L_{\text{preq}}(\mathcal{D}_N;\mathcal{L}) 
    = \sum_{s=0}^{S-1} \sum_{i=t_s}^{t_{s+1}-1} -\log p^{(s)}(y_i\mid x_i),
\end{equation}

\section{Intermediate Models and the Compression Envelope}
\label{appendix:intermediate_models}

The construction of the PCL curve in \cref{subsec:desc_length_estimation} yields, for each feature type, a single fully-converged predictor $p_N$ with complexity $L_{\mathcal{L}}(p_N)$ and asymptotic loss $\ell_{N} = \mathbb{E}_{(x,y)\sim p^*}[-\log p_N(y\mid x)]$. However, behind each fully-converged predictor, lies a whole family of intermediate predictors that rely on \textit{some} partial information about the feature (e.g. only the low-frequency geometric patterns of the digit feature). In practice, nothing prevents the learner from selecting one of those intermediate predictors instead of the fully-converged one. In fact, for the picture to be complete, we need to consider these intermediate predictors as potential candidates for compression. To this end, we derive a set of intermediate models by truncating the PCL curve for exponentially large AUC. 

Let $\ell(n)$ denote the prequential test loss of a feature-specific predictor trained on $n$ samples, and let $\ell_{\text{orig}}(n)$ denote its cross-entropy on the original distribution at training size $n$.  For a truncation point $N_t \leq N$, we define the intermediate model $p_{N_t}$ with:
\begin{itemize}
    \item \textbf{Model complexity} (fixed-cost):
    \[
        L_{\mathcal{L}}(p_{N_t}) \;\approx\; \int_{}^{N_t} \!\big[\ell(n) - \ell(N_t)\big]\,dn
    \]
    \item \textbf{Compression rate} (variable-cost): $\ell_{\text{orig}}(N_t)$.
\end{itemize}
As we approach the fully-converged model, the complexity increases while the compression rate decreases, and  as $N_t \to N$, we recover the fully-converged model. Each intermediate model defines an affine compression line $\mathcal{J}(p_{N_t}, N) = L_{\mathcal{L}}(p_{N_t}) + N \cdot \ell_{\text{orig}}(N_t)$, and the MDL-predicted feature at dataset size $N$ is determined by the \emph{lower envelope} over all intermediate models across all feature types:
\[
    \hat{p}_N \;=\; \underset{p_{N_t}^{(f)},\; f \in \mathcal{F}}{\arg\min}\;\Big[ L_{\mathcal{L}}(p_{N_t}^{(f)}) + N \cdot \ell_{\text{orig}}^{(f)}(N_t) \Big]
\]
where $\mathcal{F}$ indexes the set of candidate feature types. The feature-type transitions reported in our experiments correspond to the points where this envelope switches between models derived from different feature types.

\section{Task Definition and Sample Generation}
\label{appendix:task_details}

\paragraph{Benchmark design:}

Given a handwritten digit image $x$, the task consists in predicting whether the digit value is $d \geq 5$ (class $y=1$) or $d < 5$ (class $y=0$). We use the EMNIST digits dataset, which extends MNIST with a total of $280,000$ samples. Prequential code length being computationally very expensive to evaluate,  we intentionally keep the visual task simple (binary classification on $32 \times 32$ images) due to limited compute resources.


Each sample $(x, y, e)$ is generated according to this procedure:
\begin{enumerate}
    \item \textbf{Binary label assignment (robust feature):} Given the original digit value $d \in \{0, \ldots, 9\}$, the label is assigned as $y = \mathbb{1}[d \geq 5] \oplus \text{Bernoulli}(p_{\text{flip}})$, where $\oplus$ denotes the XOR operation. With probability $p_{\text{flip}}$, the label is flipped, introducing noise in the digit-label relationship. The digit shape is therefore the causal determinant of the label, and $p_{\text{flip}}$ controls how predictive this robust feature is.
    \item \textbf{Environment assignment:} Each sample is assigned to one of two environments using $e \sim \text{Bernoulli}(p_{e})$, where $p_{e}$ controls the environment proportion. 
    \item \textbf{Color assignment (spurious feature):} The digit is colorized based on environment and label. In \textbf{Environment 0}, $y=0 \rightarrow$ green and $y=1 \rightarrow$ red. In \textbf{Environment 1}, $y=0 \rightarrow$ red and $y=1 \rightarrow$ green. The color thus creates a low-complexity spurious correlation with $y$; its predictive strength is governed by the environment imbalance $p_{e}$. When $p_e = 0.5$, the two environments are balanced and the marginal color--label correlation vanishes. The further $p_e$ is from $0.5$, the stronger the environment imbalance, and the stronger the marginal spurious correlation between color/digit becomes. Thus, $p_e$ controls the predictiveness of the color feature.
    \item \textbf{Watermark assignment (complex environmental cue):} A binary watermark pattern is embedded in the rightmost pixel column of each image, encoding the environment through two non-overlapping banks of $K$ random patterns (one per environment). Exploiting this feature requires memorizing all $2K$ patterns, so the bank size $K$ serves as a direct control knob on the complexity of this environmental signal. Details of the watermark generation are given below.
\end{enumerate}

\paragraph{Watermark generation}
We pre-generate two watermark banks $\mathcal{B}_0$ and $\mathcal{B}_1$, each containing $B$ unique binary vectors of length $b$ bits (where $b=32$ matches the image height). The two banks are guaranteed to have no overlap, i.e.\ $\mathcal{B}_0 \cap \mathcal{B}_1 = \emptyset$. Each pattern is sampled uniformly at random from $\{0,1\}^b$, with uniqueness enforced via rejection sampling. For each sample with environment $e \in \{0,1\}$, a watermark is drawn uniformly at random from the corresponding bank: $\mathbf{b} \sim \text{Uniform}(\mathcal{B}_e)$, and the rightmost pixel column of the image is set to $\mathbf{b}$.

The \textbf{bank size} $B$ directly controls the complexity of the watermark feature: a model exploiting watermarks must memorize all $2B$ distinct patterns and their environment assignments. Small values of $B$ (e.g., $B=2$) yield a simple spurious feature that is easy to memorize, while large values (e.g., $B=50$) produce a complex feature requiring substantially more model capacity to exploit.

\subsection{Task examples}

\begin{figure}[h]
    \centering
    \includegraphics[width=0.9\textwidth]{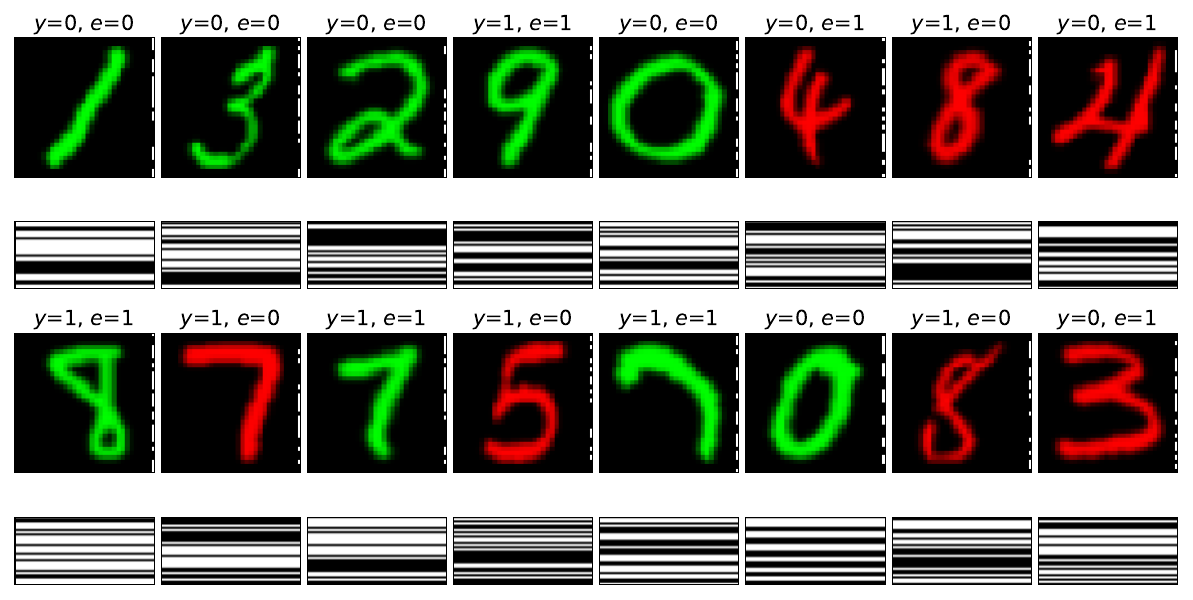}
    
    \caption{Example samples from the EMNIST-based benchmark we curated, featuring the three types of features we presented in \cref{subsec:robust}: the digit shape (robust feature), the color (simple spurious feature) and the watermark (complex environmental-cue). The label is determined by the digit value, while the color and watermark are correlated with the label in an environment-dependent way.
    }
    \label{fig:task_samples}
\end{figure}
\section{Qualitative Study: Experimental Details}
\label{appendix:experiment_details}

The side-by-side comparison in \cref{fig:qualitative_study} considers different configurations of Color-MNIST:

\paragraph{Scenario A} $p_e = 0.25$, $p_{\text{flip}} = 0$, no watermark. The color feature is simpler but less predictive than the digit feature. The theory predicts that color dominates at small $N$ before the learner switches to digit as the suboptimal compression rate becomes prohibitive.

\paragraph{Scenario B} $p_e = 0.5$, $p_{\text{flip}} = 0.15$, watermark bank size $K = 50$. Balanced environments eliminate the marginal color-label correlation. The digit feature is simpler but noisy (85\% ceiling). The theory predicts digit dominance at intermediate $N$, with a transition to the watermark-based model at large $N$.

\section{Training Procedure and Hyperparameters}
\label{appendix:training_hyperparameters}

We train neural networks using standard Empirical Risk Minimization (ERM) with AdamW. ERM aims to minimize the cross-entropy loss on the training data, which corresponds to minimizing the negative log-likelihood of the model's predictions. For all experiments, we use the following training configuration (see \cref{tab:hyperparameters} for a summary):

\paragraph{Architecture} Our predictive model is a simple Multi-Layer Perceptron (MLP) with a featurizer consisting of 2 hidden layers of dimension 256, followed by a linear classification head. The featurizer processes flattened $32 \times 32$ RGB images into $3072$-dimensional vectors, uses ReLU activations, and Xavier (Glorot) uniform initialization. We apply no dropout. While our theory is architecture-agnostic, we restrict this study to simple MLPs and toy datasets due to the extreme computational cost of prequential code length estimation (each point in the PCL curve requires training a model from scratch until convergence)

\paragraph{Optimization} We use AdamW optimizer with learning rate $\eta = 10^{-3}$ and $L_2$ weight decay (regularization) $\lambda = 10^{-4}$. The batch size is set to $B = 64$ for all experiments. We train until convergence using early stopping with a patience of 3 epochs and a minimum improvement threshold of $\Delta = 5 \times 10^{-4}$.

\paragraph{Data Sampling} For each dataset size $N$, we train the model on a random subset of size $N$ sampled from the full training set. To reduce variance in estimates for small dataset sizes ($N \leq 500$), we perform 10 independent runs with different random subsets and seeds, averaging the resulting metrics over the runs. For larger datasets ($N > 500$), we use 3 independent runs per dataset size.

\paragraph{Implementation} All experiments are implemented in Python 3.12 using PyTorch 2.9. Training is performed on NVIDIA GPUs with CUDA 11.8. The source code is available in the \texttt{complicity} repository.

\begin{table}[h]
\centering
\caption{Hyperparameters for the Watermarked CMNIST Experiments}
\begin{tabular}{lcll}
\hline
\textbf{Parameter} & \textbf{Symbol} & \textbf{Description} & \textbf{Values} \\ \hline
\multicolumn{4}{l}{\textit{Environment \& Spurious Correlation}} \\
\texttt{flip\_prob} & $p_{\text{flip}}$ & Label noise probability & $\{0, 0.01, \ldots, 0.1\}$ \\
\texttt{spur\_prob} & $p_{e}$ & Relative proportion of environments & $\{0, 0.02, \ldots, 0.2\}$ or $0.5$ \\
\texttt{uninformative\_majority} & --- & Randomize color in majority env & \texttt{True}, \texttt{False} \\ \hline
\multicolumn{4}{l}{\textit{Watermark Configuration}} \\
\texttt{watermark\_bank\_size} & $K$ & Distinct patterns per environment & $1$ to $100$ (log-spaced) \\
\texttt{watermark\_bits} & $b$ & Length of each binary pattern & $32$ \\
\texttt{random\_watermark} & --- & Use random (non-informative) watermark & \texttt{True}, \texttt{False} \\ \hline
\multicolumn{4}{l}{\textit{Image Configuration}} \\
\texttt{input\_size} & --- & Image resolution & $32 \times 32$ \\
\texttt{Digit} & --- & Remove color information & \texttt{True}, \texttt{False} \\
\texttt{noise\_digit} & --- & Replace digit with Gaussian noise & \texttt{True}, \texttt{False} \\ \hline
\multicolumn{4}{l}{\textit{Model Architecture (MLP)}} \\
\texttt{n\_outputs} & $h$ & Hidden layer dimension & $256$ \\
\texttt{n\_layers} & $L$ & Number of hidden layers & $2$ \\
\texttt{mlp\_dropout} & $p_{\text{drop}}$ & Dropout probability & $0.0$ \\ \hline
\multicolumn{4}{l}{\textit{Optimization}} \\
\texttt{optimizer} & --- & Optimization algorithm & AdamW \\
\texttt{lr} & $\eta$ & Learning rate & $10^{-3}$ \\
\texttt{weight\_decay} & $\lambda$ & L2 regularization & $10^{-4}$ \\
\texttt{batch\_size} & $B$ & Mini-batch size & $64$ 
\end{tabular}
\label{tab:hyperparameters}
\end{table}

\end{document}